\documentclass[letterpaper]{article} 
\usepackage{aaai23}  
\usepackage{times}  
\usepackage{helvet}  
\usepackage{courier}  
\usepackage[hyphens]{url}  
\usepackage{graphicx} 
\usepackage{natbib}  
\usepackage{caption} 
\usepackage[ruled, linesnumbered]{algorithm2e}
\usepackage{algorithmic}

\usepackage{amsmath}
\usepackage{bm}
\usepackage{mathrsfs}
\usepackage{amsthm}
\usepackage{amsfonts}
\usepackage{subfigure}
\usepackage{multirow}
\usepackage{url}
\usepackage{array}
\usepackage{amssymb}
\usepackage{booktabs} 

\newcommand{\note}[1]{{{     \emph{\underline{#1} }    }}}

\title{Self-Supervised Continual Graph Learning in Adaptive Riemannian Spaces}
\author{
    Li Sun\textsuperscript{\rm 1}\thanks{Corresponding Authors: Li Sun and Hao Peng},
    Junda Ye\textsuperscript{\rm 2},
    Hao Peng\textsuperscript{\rm 3}$^*$,
    Feiyang Wang\textsuperscript{\rm 2},
    Philip S. Yu\textsuperscript{\rm 4}
}
\affiliations{
    \textsuperscript{\rm 1}School of Control and Computer Engineering, North China Electric Power University, Beijing 102206, China\\
    \textsuperscript{\rm 2}School of Computer Science, Beijing University of Posts and Telecommunications, Beijing 100876, China\\
    \textsuperscript{\rm 3}Beijing Advanced Innovation Center for Big Data and Brain Computing, Beihang University, Beijing 100191, China\\
    \textsuperscript{\rm 4}Department of Computer Science, University of Illinois at Chicago, IL, USA\\
    ccesunli@ncepu.edu.cn; \{jundaye, fywang\}@bupt.edu.cn; penghao@buaa.edu.cn; psyu@uic.edu

}

\begin{document}

\maketitle


\begin{abstract}
Continual graph learning routinely finds its role in a variety of real-world applications where the graph data with different tasks come sequentially.
Despite the success of prior works, it still faces great challenges.
On the one hand, existing methods work with the zero-curvature Euclidean space, and largely ignore the fact that curvature varies over the coming graph sequence.
On the other hand, continual learners in the literature rely on abundant labels, but labeling graph in practice is particularly hard especially for the  continuously emerging graphs on-the-fly.
To address the aforementioned challenges, we propose to explore a challenging yet practical problem, \emph{the self-supervised continual graph learning in adaptive Riemannian  spaces}. 
In this paper, we propose a novel  self-supervised \underline{Rie}mannian \underline{Gra}ph \underline{C}ontinual  L\underline{e}arner (\textbf{RieGrace}).
In RieGrace, we first design an Adaptive Riemannian GCN (AdaRGCN), a unified GCN coupled with a neural curvature adapter, so that Riemannian space is shaped by the learnt curvature  \emph{adaptive to each graph}.
Then, we present a \emph{Label-free Lorentz Distillation} approach, in which we create teacher-student AdaRGCN for the graph sequence.
The student successively performs intra-distillation from itself and inter-distillation from the teacher so as to consolidate knowledge without catastrophic forgetting.
In particular, we propose a theoretically grounded Generalized Lorentz Projection for the contrastive distillation in Riemannian space.
Extensive experiments on the benchmark datasets show the superiority of RieGrace, and additionally, we investigate on how curvature changes over the graph sequence.
\end{abstract}

\section{Introduction}
Continual graph learning is emerging as a hot research topic which successively learns from a graph sequence with different tasks \cite{DBLP:journals/corr/abs-2202-10688}. 
In general, it aims at gradually learning new knowledge without \emph{catastrophic forgetting} the old ones across sequentially coming tasks.
Centered around fighting with forgetting, a series of methods \cite{DBLP:conf/ijcai/KimYK22,DBLP:conf/ijcnn/GalkeFZS21} have been proposed recently.
Despite the success of prior works, continual graph learning still faces tremendous challenges. 

\emph{\textbf{Challenge 1}: An adaptive Riemannian representation space.}
To the best of our knowledge, existing methods work with Euclidean space, the zero-curvature Riemannian space \cite{DBLP:journals/corr/abs-2111-15422,DBLP:conf/aaai/0002C21,DBLP:conf/cikm/WangSWW20}.
However, in continual graph learning, the curvature of a graph remains unknown until its arrival. 
In particular, the negatively curved Riemannian space, hyperbolic space, is well-suited for graphs presenting hierarchical patterns or tree-like structures \cite{krioukov2010hyperbolic,nickel2017poincare}.
The underlying geometry shifts to be positively curved, hyperspherical space, when cyclical patterns (e.g., triangles or cliques) become dominant \cite{BachmannBG20}.
Even more challenging,  the curvature usually varies over the coming graph sequence as shown in the case study.
Thus, it calls for a smart graph encoder in the Riemannian space with \emph{adaptive curvature} for each coming graph successively.

\emph{\textbf{Challenge 2}: Continual graph learning without supervision.}
Existing continual graph learners \cite{MCGL-NAS,FGN} are trained in the supervised fashion, and thereby rely on abundant labels for each learning task.
Labeling graphs requires either manual annotation or paying for permission in practice. It is particularly hard and even impossible when graphs are continuously emerging on-the-fly.
In this case, \emph{self-supervised learning} is indeed appealing, so that we can acquire knowledge from the unlabeled data themselves.
Though self-supervised learning on graphs is being extensively studied \cite{VelickovicFHLBH19,QiuCDZYDWT20,DBLP:conf/aaai/YinWHXZ22}, existing methods are trained offline. That is, they are not applicable for continual graph learning, and naive application results in catastrophic forgetting in the successive learning process  \cite{DBLP:journals/pami/LangeAMPJLST22,DBLP:conf/nips/KeLMXS21}.
Unfortunately, self-supervised continual graph learning is surprisingly under-investigated in the literature.

Consequently, it is vital to explore how to learn and memorize knowledge free of labels for continual graph learning in adaptive Riemannian spaces. 
Thus, we propose the challenging yet practical problem of \emph{self-supervised continual graph learning in adaptive Riemannian spaces}. 

In this paper, we propose a novel self-supervised \underline{Rie}mannian \underline{Gra}ph \underline{C}ontinual  L\underline{e}arner (\textbf{RieGrace}).
To address the first challenge, we design an Adaptive Riemannian GCN (AdaRGCN), which is able to \emph{shift among any hyperbolic or hyperspherical space adaptive to each graph}.
In AdaRGCN, we formulate a unified Riemannian graph convolutional network (RGCN) of arbitrary curvature, and design a  CurvNet inspired by Forman-Ricci curvature in Riemannian geometry.
CurvNet is a neural module in charge of curvature adaptation, so that we induce a Riemannian space shaped by the curvature learnt from the task graph.
To address the second challenge, we propose a novel \emph{label-free Lorentz distillation approach to consolidate knowledge without catastrophic forgetting}.
Specifically, we create teacher-student AdaRGCN for the graph sequence.
When receiving a new graph, the student is created from the teacher.
The student distills from the intermedian layer of itself to acquire knowledge of current graph (intra-distillation), and in the meanwhile, 
distills from the teacher to preserve the past knowledge (inter-distillation).
In our approach, we propose to consolidate knowledge via contrastive distillation, but it is particularly challenging to contrast between different Riemannian spaces. 
To bridge this gap, we formulate a novel \emph{Generalized Lorentz Projection} (GLP).
We prove GLP is closed on Riemannian spaces, and show its relationship to the well-known Lorentz transformation.

In short, noteworthy contributions are summarized below:
\begin{itemize}
  \item \emph{Problem}. We propose the problem of self-supervised continual graph learning in adaptive Riemannian spaces, which  is the first attempt,  to the best of our knowledge,  to study continual graph learning in non-Euclidean space.
  \item \emph{Methodology}. We present a novel RieGrace, where we design a unified RGCN with CurvNet to 
  shift curvature among hyperbolic or hyperspherical spaces adaptive to each graph,
  and propose the Label-free Lorentz Distillation with GLP for self-supervised continual learning.
\item \emph{Experiments}. Extensive experiments on the benchmark datasets show that RieGrace even outperforms the state-of-the-arts supervised methods, and the case study gives further insight on the curvature over the graph sequence with the notion of embedding distortion.
\end{itemize}

\vspace{-0.1in}
\section{Preliminaries}
\vspace{-0.02in}
In this section, we first introduce the fundamentals of Riemannian geometry necessary to understand this work, and then formulate the studied problem, \emph{self-supervised continual graph learning in general Riemannian space}.
In short, we are interested in how to learn an encoder $\mathbf \Phi$ that is able to sequentially learn on coming graphs $G_1, \dots, G_T$  in adaptive Riemannian spaces without external supervision.

\vspace{-0.09in}

\subsection{Riemannian Geometry}

\vspace{-0.03in}
\subsubsection{Riemannian Manifold.} 
A Riemannian manifold $(\mathcal M, g)$ is a smooth manifold $\mathcal M$ equipped with a Riemannian metric $g$.
Each point $\mathbf x$ on the manifold is associated with a \emph{tangent space} $\mathcal T_\mathbf x\mathcal M$ that looks like Euclidean.
The Riemannian metric $g$ is the collection of inner products at each point  $\mathbf x \in \mathcal M$ regarding its tangent space.
For $\mathbf x \in \mathcal M$, 
the  \emph{exponential map} at $\mathbf x$, $exp_\mathbf x(\mathbf v): \mathcal T_\mathbf x\mathcal M \to \mathcal M$, 
projects the vector of the tangent space at $\mathbf x$ onto the manifold, 
and the \emph{logarithmic map} $log_\mathbf x(\mathbf y): \mathcal M \to \mathcal T_\mathbf x\mathcal M $ is the inverse operator.

\vspace{-0.03in}
\subsubsection{Curvature.} In Riemannian geometry, the \emph{curvature} is the notion to measure how a smooth manifold deviates from being flat.
If the curvature is uniformly distributed, the manifold $\mathcal M$ is called the space of constant curvature $\kappa$. 
In particular, the space is \emph{hyperspherical} $\mathbb S$  with  $\kappa>0$ when it is positively curved, and \emph{hyperbolic} $\mathbb H$ with $\kappa<0$ when negatively curved.
Euclidean space is flat space with $\kappa=0$, and can be considered as a special case in Riemannian geometry.


\vspace{-0.05in}
\subsection{Problem Formulation}
In the continual graph learning, we will receive a sequence of disjoint tasks $\mathcal T=\{\mathcal T_1, \dots, \mathcal T_t, \dots, \mathcal T_T\}$, and each task is defined on a graph $G=\{\mathcal V, \mathcal E\}$, where
$\mathcal V=\{v_1, \cdots, v_N\}$ is the node set, and $\mathcal E=\{(v_i, v_j)\} \subset \mathcal V \times \mathcal V$ is the edge set.
Each node $v_i$ is associated with node feature $\mathbf x_i$ and  a category $y_i \in \mathcal Y_k$, where $\mathcal Y_k$ is the label set of $k$ categories.
\vspace{-0.02in}
\newtheorem*{def1}{Definition 1 (Graph Sequence)} 
\begin{def1}
The sequence of tasks in graph continual learning is described as a graph sequence $\mathcal G=\{G_1, \dots, G_T\}$, and each graph $G_t$ corresponds to a task $\mathcal T_t$. Each task contains a training node set $\mathcal V_t^{tr}$ and a testing node set $\mathcal V_t^{te}$ with node features $X_t^{tr}$ and $X_t^{te}$.
\vspace{-0.02in}
\end{def1}

In this paper, we study the task-incremental learning in \emph{adaptive Riemannian space} whose curvature is able to successively match each task graph.
When a new graph arrives, the learnt parameters are memorized but historical graphs are dropped, and additionally, no labels are provided in the learning process.
We give the formal definition as follows:
\vspace{-0.02in}
\newtheorem*{prob}{Definition 2 (Self-Supervised Continual Graph Learning in Adaptive Riemannian Space)} 
\begin{prob}
Given a graph sequence $\mathcal G$ with tasks $\mathcal T$, we aim at learning an encoder $\mathbf \Phi: v \to  \mathbf h \in \mathcal M^{d,k}$ in absence of labels in adaptive Riemannian space,
so that the encoder is able to continuously consolidate the knowledge for current task without catastrophically forgetting the knowledge for previous ones.
\vspace{-0.02in}
\end{prob}
Essentially different from the continual graph learners of prior works, we study with a more challenging yet practical setting:  i) rather than Euclidean space, the encoder $\mathbf \Phi$ works with an adaptive Riemannian space suitable for each task, and ii)  is able to learn and memorize knowledge without labels for continuously emerging graphs on-the-fly.

\begin{figure*}
\centering
       \vspace{-0.15in}
    \includegraphics[width=0.97\linewidth]{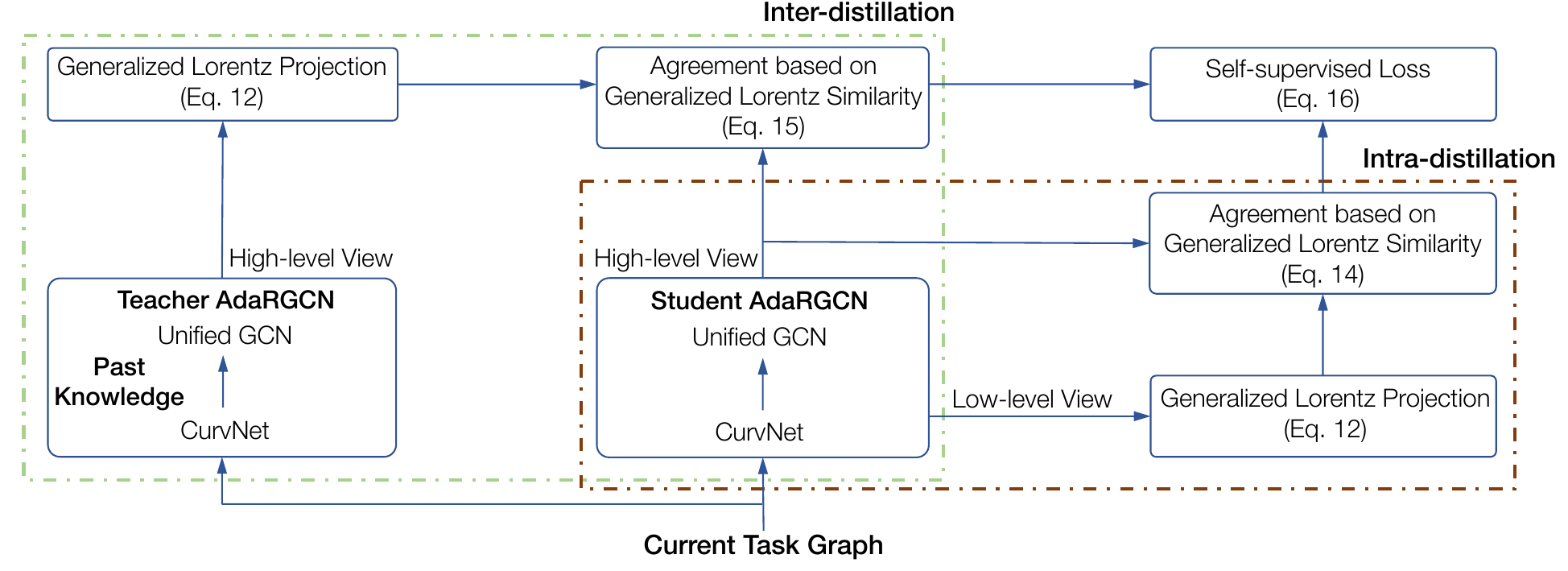}
    \vspace{-0.13in}
    \caption{Overall architecture of \textbf{RieGrace}. 
    We design the AdaRGCN which successively adapts its curvature for current task graph with CurvNet, and propose Label-free Lorentz Distillation for continual graph learning. In each learning session, i) the student is created from the teacher with the same architecture,
    ii) jointly performs intra-distillation from itself and inter-distillation from the teacher with GLP to consolidate knowledge, and iii) becomes the teacher for the next learning session. 
    }
    \label{illu}
        \vspace{-0.2in}
\end{figure*}

\vspace{-0.03in}
\section{Methodology}
\vspace{-0.03in}
To address this problem, we propose a novel Self-supervised \underline{Rie}mannian \underline{Gra}ph \underline{C}ontinual  L\underline{e}arner (\textbf{RieGrace})
We illustrate the overall architecture of RieGrace in Figure 1. 
In the nutshell, we first design a unified graph convolutional network (AdaRGCN) on the Riemannian manifold shaped by the learnt curvature \emph{adaptive to each coming graph}.
Then, we propose a \emph{label-free Lorentz distillation approach} to consolidate knowledge without catastrophic forgetting.

\vspace{-0.07in}
\subsubsection{Representation Space.} First of all, we introduce the Riemannian manifolds we use in this paper before we construct RieGrace on them. We opt for the hyperboloid (Lorentz) model for hyperbolic space and the corresponding hypersphere model for hyperspherical space with the unified formalism, owing to the numerical stability and closed form expressions \cite{HGNN}.


Formally, we have a  $d$-dimensional manifold of curvature $\kappa$,
$\mathcal M^{d, \kappa}=\{
\mathbf x \in \mathbb R^{d+1} | \  \langle \mathbf{x}, \mathbf{x} \rangle_\kappa= \frac{1}{\kappa}\} $
with $\kappa\neq 0$,
whose \emph{origin} is denoted as $\mathcal O = (|\kappa|^{-\frac{1}{2}}, 0, \cdots, 0) \in \mathcal M^{d,\kappa}$.
The curvature-aware inner product $\langle \cdot, \cdot \rangle_\kappa$ is defined as 
\vspace{-0.07in}
\begin{equation}
\langle \mathbf{x}, \mathbf{y} \rangle_\kappa=\mathbf{x}^{\top} \operatorname{diag}(sign(\kappa),1, \cdots, 1) \mathbf{y},
\vspace{-0.07in}
\label{manifold}
\end{equation}
and thus the tangent space at $\mathbf x$ is given as $\mathcal T_\mathbf x\mathcal M^{d,\kappa}=\{\mathbf v \in \mathbb R^{d+1} | \  \langle \mathbf{v}, \mathbf{x} \rangle_\kappa= 0\}$.
In particular, 
for the positive curvature, $\mathcal M^{d, \kappa}$ is the hypersphere model $\mathbb S^{d, \kappa}$ and $\langle \cdot, \cdot \rangle_\kappa$ is the the standard inner product on $\mathbb R^{d+1}$.
For the negative curvature, $\mathcal M^{d, \kappa}$ is the hyperboloid model $\mathbb H^{d, \kappa}$ and $\langle \cdot, \cdot \rangle_\kappa$ is the Minkowski inner product.
The operators with the unified formalism on $\mathcal M^{d, \kappa}$ is summarized in Table \ref{tab:ops}, where 
$v=\sqrt{|\kappa|}\|\mathbf{v}\|_{\kappa}$, $\beta = \kappa\langle\mathbf{x}, \mathbf{y}\rangle_{\kappa}$ and $\| \mathbf{v} \|_{\kappa}^2=\langle \mathbf{v}, \mathbf{v} \rangle_\kappa$ for $\mathbf v \in \mathcal T_\mathbf x\mathcal M^{d, \kappa}$.
We utilize the curvature-aware trigonometric function the same as \citet{SkopekGB20}.

\begin{table}
\centering
\begin{tabular}{|l|c|}
\hline
\textbf{Operator}  & \textbf{Unified formalism in $\mathcal M^{d, \kappa}$}\\
\hline
Distance Metric
&
$
d_{\mathcal M}(\mathbf{x}, \mathbf{y})=\frac{1}{\sqrt{|\kappa|}} \cos^{-1}_{\kappa}\left( \kappa \langle\mathbf{x}, \mathbf{y}\rangle_{\kappa}\right)
$\\
\hline
Exponential Map & 
$
exp _{\mathbf{x}}^{\kappa}(\mathbf{v})=\cos_{\kappa}\left(\alpha \right) \mathbf{x}+ \frac{\sin_{\kappa}\left(\alpha \right)}{\alpha}\mathbf{v}
$
\\
Logarithmic Map & 
$
log _{\mathbf{x}}^{\kappa}(\mathbf{y})=\frac{\cos^{-1}_{\kappa} \left(\beta \right)}{\sin _{\kappa}\left(\cos^{-1}_{\kappa}\left( \beta \right)\right)}\left(\mathbf{y}-\beta \mathbf{x}\right)
$ 
\\
\hline
Scalar Multiply &
$
r \otimes_{\kappa} \mathbf{x}=exp _{\mathcal O}^{\kappa}\left(r \ log_{\mathcal O}^{\kappa}(\mathbf{x})\right) 
$\\
\hline
\end{tabular}
\vspace{-0.1in}
\caption{Curvature-aware operations in manifold $\mathcal M^{d, \kappa}$.}
\vspace{-0.2in}
\label{tab:ops}
\end{table}

\vspace{-0.09in}
\subsection{Adaptive Riemannian GCN}
\vspace{-0.03in}

Recall that the curvature of task graph remains unknown until its arrival. 
We propose an adaptive Riemannian GCN (AdaRGCN), a unified GCN of arbitrary curvature coupled with a CurvNet, a neural module for curvature adaptation. 
AdaRGCN shifts among hyperbolic and hyperspherical spaces accordingly to match the geometric  pattern of each graph, essentially distinguishing itself from prior works. 

\vspace{-0.05in}
\subsubsection{Unified GCN of Arbitrary Curvature.}

Recent studies in Riemannian graph learning mainly focus on the design of GCNs in manifold $\mathcal M^{d,\kappa}$ with negative curvatures (hyperboloid model), but the unified GCN of arbitrary curvature has rarely been touched yet.
To bridge this gap, we propose a unified GCN of arbitrary curvature, generalizing from the zero-curvature GAT \cite{velickovic2018graph}. 
Specifically, we introduce the operators with unified formalism on $\mathcal M^{d,\kappa}$.

Feature transformation is a basic operation in neural network. For $\mathbf{h} \in \mathcal M^{d,\kappa}$, 
we perform the transformation via the $\kappa$-left-multiplication $\boxtimes_\kappa$ defined by $exp _{\mathcal O}^{\kappa}(\cdot)$ and $log _{\mathcal O}^{\kappa}(\cdot)$, 
\vspace{-0.05in}
\begin{equation}
\mathbf W  \boxtimes_{\kappa} \mathbf{h}= exp _{\mathcal O}^{\kappa}\left(\left[ \ 0 \ \| \ \mathbf W \ log_{\mathcal O}^{\kappa}(\mathbf{h})_{[1:d]} \right]\right),
\vspace{-0.05in}
\label{ft}
\end{equation}
where $\mathbf{W} \in \mathbb R^{d' \times d}$ is weight matrix, and $[ \cdot \| \cdot ]$ denotes concatenation. Note that, $[log _{\mathcal O}^{\kappa}(\mathbf{h})]_0=0$ holds, $\forall \mathbf{h} \in \mathcal M^{d,\kappa}$.

The advantage of Eq. (\ref{ft}) is that \emph{logarithmically mapped vector lies in the tangent space $\mathcal T_\mathcal O\mathcal M^{d,\kappa}$ for any $\mathbf W$ so that we can utilize $exp _{\mathcal O}^{\kappa}(\cdot)$ safely}, which is not guaranteed in direct combination formalism of $exp _{\mathcal O}^{\kappa}( \mathbf Wlog _{\mathcal O}^{\kappa}(\mathbf{h}))$. Similarly, we give the formulation of applying function $f(\cdot)$,
\vspace{-0.035in}
\begin{equation}
f_{\kappa}(\mathbf{h})= exp _{\mathcal O}^{\kappa}\left(\left[ \  0\  \|  \ f(log_{\mathcal O}^{\kappa}(\mathbf{h}))_{[1:d]} \right]\right).
\vspace{-0.035in}
\end{equation}

Neighborhood aggregation  is a weighted \emph{arithmetic mean} and also the \emph{geometric centroid} of the neighborhood features essentially \cite{DBLP:conf/icml/WuSZFYW19}.
Fr\'{e}chet mean follows this meaning in Riemannian space, but unfortunately does not have a closed form solution \cite{DBLP:conf/icml/LawLSZ19}.
Alternatively, we define neighborhood aggregation as the geometric centroid of squared distance, in spirit of Fr\'{e}chet mean, to enjoy both mathematical meaning and efficiency.
Given a set of neighborhood features $\mathbf{h}_{j}\in \mathcal M^{d,\kappa}$ centered around $v_i$,
the closed form aggregation  is derived as follows:
\vspace{-0.07in}
\begin{equation}
\resizebox{0.905\hsize}{!}{$
 AGG_\kappa(\{\mathbf{h}_{j}, \nu_{ij} \}_i)=  \frac{1}{\sqrt{|\kappa|} } \sum\nolimits_{j \in \bar{\mathcal N}_i}\frac{\nu_{ij}\mathbf{h}_{j}}{\left| ||\sum\nolimits_{j \in \bar{\mathcal N}_i} \nu_{ij}\mathbf{h}_{j}||_\kappa \right|},
 $}
 \label{agg}
 \vspace{-0.07in}
 \end{equation}
where $\bar{\mathcal N}_i$ is the neighborhood of $v_i$ including itself, and $\nu_{ij}$ is the attention weight.

Different from \citet{DBLP:conf/icml/LawLSZ19,ZhangWSLS21}, we generalize the centroid from hyperbolic space to the Riemannian space of arbitrary curvature $\mathcal M^{d,\kappa}$,
and show its connection to gyromidpoint of  $\kappa$-sterographical model theoretically.
Now, we prove that \emph{arithmetic mean} in Eq. (\ref{agg}) is the closed form expression of the \emph{geometric centroid}.
\newtheorem*{prop1}{Proposition 1} 
\begin{prop1}
Given a set of points $\mathbf{h}_{j}\in \mathcal M^{d,\kappa}$ each attached with a weight $\nu_{ij}$, $j \in \Omega$,
the centroid  of squared distance $ \mathbf{c}$ in the manifold is given as minimization problem:
\vspace{-0.1in}
\begin{equation}
\min _{ \mathbf c \in \mathcal{M}^{d, \kappa}} \ \  \sum\nolimits_{j \in \Omega} \nu_{i j} d_{\mathcal M}^{\ 2}\left(\mathbf{h}_{j}, \mathbf{c}\right),
\vspace{-0.07in}
\end{equation}
Eq. (\ref{agg}) is the closed form solution, $\mathbf c=AGG_\kappa(\{\mathbf{h}_{j}, \nu_{ij} \}_i)$. 
\end{prop1}
\vspace{-0.1in}
\begin{proof}
We have 
$
\mathbf{c} = \arg \min _{ \mathbf c \in \mathcal{M}^{d, \kappa}} \sum_{j \in \Omega} \nu_{i j} d_{\kappa}^{\ 2}\left(\mathbf{h}_{j}, \mathbf{c}\right)
$,
and $ \mathbf{c}$ is in the manifold $\mathbf c \in \mathcal{M}^{d, \kappa}$, i.e., $\langle \mathbf{c}, \mathbf{c} \rangle_\kappa=\frac{1}{\kappa}$.
Please refer to the Appendix for the details.
\end{proof}
\vspace{-0.1in}


Attention mechanism is equipped for neighborhood aggregation as the importance of neighbor nodes are usually different.
We study the importance between a neighbor $v_j$ and center node $v_i$ by an attention function in tangent space, 
\vspace{-0.1in}
\begin{equation}
ATT_\kappa(\mathbf{x}_i, \mathbf{x}_j, \mathbf{\theta})=\boldsymbol{\theta}^\top \left[ log_{\mathcal O}^{\kappa}(\mathbf{x}_i) ||  log_{\mathcal O}^{\kappa}(\mathbf{x}_j) \right],
\end{equation}
parameterized by $\boldsymbol{\theta}$, and then attention weight is given by 
$\nu_{ij}=Softmax_{j \in \mathcal N_i}(ATT_\kappa(\mathbf{x}_i, \mathbf{x}_j, \boldsymbol{\theta}))$.

We formulate the convolutional layer on  $\mathcal{M}^{d, \kappa}$ with proposed operators. The message passing in the $l^{th}$ layer is 
\vspace{-0.07in}
\begin{equation}
\mathbf{h}_i^{(l)}=\delta_\kappa\left(AGG_\kappa(\{\mathbf{x}_{j}, \nu_{ij} \}_i) \right), \mathbf{x}_i=\mathbf{W} \boxtimes_{\kappa} \mathbf{h}_i^{(l-1)},
\vspace{-0.05in}
\end{equation}
where $\delta_\kappa(\cdot)$ is the nonlinearity.
Consequently, we build the unified GCN by stacking multiple convolutional layers, 
and its curvature is adaptively learnt for each task graph with a novel neural module designed as follows.

\vspace{-0.07in}
\subsubsection{Curvature Adaptation.}
We aim to learn the curvature of any graph with a function $f: G \to \kappa$, so that the Riemannian space is able to successively match the geometric pattern of the task graph.
To this end, we design a simple yet effective network, named CurvNet, based on the notion of Forman-Ricci curvature in  Riemannian geometry.

\emph{Theory on Graph Curvature}: 
Forman-Ricci curvature defines the curvature for an edge $(v_i, v_j)$, and 
\citet{DBLP:journals/compnet/WeberSJ17} give the reformulation on the neighborhoods of its two end nodes,
\vspace{-0.05in}
\begin{equation}
\resizebox{0.909\hsize}{!}{$ 
F_{ij}=w_i+w_j-\sum\nolimits_{l\in \mathcal N_i}\sqrt{\frac{\gamma_{ij}}{\gamma_{il}}}w_l-\sum\nolimits_{k\in \mathcal N_j}\sqrt{\frac{\gamma_{ij}}{\gamma_{ik}}}w_k,
$}
\vspace{-0.05in}
\end{equation}
where $w_i$ and $\gamma_{ij}$ are the weights associated with nodes and edges, respectively. $w_i$ is defined by the degree information of the nodes connecting to $v_i$, and $\gamma_{ij}=\frac{w_i}{\sqrt{w_i^2+w_j^2}}$.
According to \cite{DBLP:conf/aaai/CruceruBG21}, $v_i$'s curvature is then given by averaging $F_{ij}$ over its neighborhood.
In other words, the curvature of a node is induced by the node weights over its 2-hop neighborhood.

We propose \textbf{CurvNet}, a 2-layer graph convolutional net, to approximate the map from node weights to node curvatures. 
CurvNet aggregates and transforms the information over 2-hop neighborhood by stacking  convolutional layer,
\vspace{-0.07in}
\begin{equation}
\mathbf Z^{(l)}=\operatorname{GCN}(\mathbf Z^{(l-1)}, \mathbf M^{(l)}),
\vspace{-0.07in}
\end{equation}
twice, where $\mathbf M^{(l)}$ is the $l^{th}$ layer parameters. 
CurvNet can be built with any GCN, and we utilize \citet{kipf2016semi} in practice.
The input features are node weights defined by degree information, $\mathbf Z^{(0)}=\mathbf A\operatorname{diag}(d_1,\cdots, d_N)$. $\mathbf A$ is the adjacency matrix, and $d_i$ is the degree of $v_i$.
The graph curvature $\kappa$ is given as the mean of node curvatures \cite{DBLP:conf/aaai/CruceruBG21}, and accordingly, we readout the graph curvature by $\kappa=MeanPooling(\mathbf Z^{(2)})$.

\vspace{-0.07in}
\subsection{Label-free Lorentz Distillation}
To consolidate knowledge free of labels,
we propose the \emph{Label-free Lorentz Distillation} approach for continual graph learning, in which we create teacher-student AdaRGCN as shown in Figure 1.
In each learning session, the student acquires knowledge for current task graph $G_t$ by distilling from itself, \emph{intra-distillation}, and preserves past knowledge by distilling from the teacher, \emph{inter-distillation}. The student finished intra- and inter-distillation becomes the teacher when new task $G_{(t+1)}$  arrives, so that we successively consolidate knowledge in the graph sequence without catastrophic forgetting.

\begin{algorithm}
        \caption{\textbf{RieGrace}. Self-Supervised Continual Graph Learning in Adaptive Riemannian Spaces} 
        \KwIn{Current tack $ G_t$, Parameters learnt from previous tasks $ G_1, \cdots, G_{t-1}$ }
        \KwOut{Parameters of AdaRGCN}
\While{not converging}{   
            \tcp{Teacher-Student AdaRGCN} 
            Froze the parameters of the teacher network\;
            $\mathbf X^{t,H} \gets \text{AdaRGCN}_{teacher}$\;
            $\{\mathbf X^{s,H}, \mathbf X^{s,L}\}\gets \text{AdaRGCN}_{student}$\;
            \tcp{Label-Free Distillation(GLP)}
            \For{each node $v_i$ in $G_t$}{
                            \emph{Intra-distillation}: Learn for current task by contrasting with Eq. (\ref{self1})\; 
                            \emph{Inter-distillation}: Learn from the teacher by contrasting with Eq. (\ref{teach1})\; 
            }
            \tcp{Update Student Parameters}
            Compute gradients of the overall objective:
            \vspace{-0.05in}
            $$ \nabla_{\mathbf{\Theta}_{student}, \{\mathbf W, \mathbf b\} }\ \ \mathcal J_{intra}+ \lambda\mathcal J_{inter}. $$
                        \vspace{-0.2in}
}
\end{algorithm}

In our approach,
we propose to distill knowledge via contrastive loss in Riemannian space. 
Though knowledge distillation has been applied to video and text \cite{GuoJY23} and similar idea on graphs has been proposed in Euclidean space \cite{DBLP:conf/aaai/0006P00LZ022,DBLP:conf/iclr/TianKI20}, they CANNOT be applied to Riemannian space owing to essential distinction in geometry.
Specifically, it lacks a method to \emph{contrast between Riemannian spaces with either different dimension or different curvature} for  the distillation. 
To bridge this gap, we propose a novel formulation, \emph{Generalized Lorentz Projection.}


\vspace{-0.1in}
\subsubsection{Generalized Lorentz Projection (GLP) \& Lorentz Layer.}

We aim to contrast between  $\mathbf x \in \mathcal M^{d_1, \kappa_1}$ and $\mathbf y \in \mathcal M^{d_2, \kappa_2}$.
The obstacle is that both dimension and curvature are incomparable ($d_1 \neq d_2$, $\kappa_1 \neq \kappa_2$). 
A naive way is to use logarithmic and exponential maps with a tangent space.
However, these maps are range to infinity, and trend to suffer from stability issue \cite{DBLP:conf/acl/ChenHLZLLSZ22}.
Such shortcomings weaken its ability for the distillation, as shown in the experiment. 


Fortunately, \emph{Lorentz transformation} in the Einstein's special theory of relativity performs directly mapping between Riemannian spaces, 
which can be decomposed into a combination of Lorentz boost $\mathbf B$  and rotation $\mathbf R$ \cite{Dragon2012}.
Formally, for $\mathbf x\in \mathcal M^{d,\kappa}$,  $\mathbf B\mathbf x\in \mathcal M^{d,\kappa}$  and $\mathbf R\mathbf x\in \mathcal M^{d,\kappa}$ 
given blocked $\mathbf B, \mathbf R \in \mathbb R^{(d+1)\times(d+1)}$ with positive semi-definiteness and special orthogonality, respectively. 
Though the clean formalism is appealing, it fails to tackle our challenge:
i) The constraints on definiteness or orthogonality render the optimization problematic. 
ii) Both dimension and curvature are fixed, i.e., they cannot be changed over time.
Recently, \citet{DBLP:conf/acl/ChenHLZLLSZ22} make effort to support different dimensions, but still restricted in the same curvature. 
Indeed, it is difficult to assure closeness of the operation especially when curvatures (i.e., shape of the manifold) are different.


In this work, we propose a novel \emph{Generalized Lorentz Projection} (GLP) in spirit of  Lorentz transformation so as to map between \emph{Riemannian spaces with different dimensions or curvatures}.
To avoid the constrained optimization, 
we reformalize GLP to learn a transformation matrix $\mathbf{W}\in \mathbb R^{d_2 \times d_1}$.
The relational behind is that $\mathbf{W}$ linearly transforms both dimension and curvature with a carefully designed formulation based on a Lorentz-type multiplication.
Formally, given $\mathbf x\in \mathcal M^{d_1,\kappa_1}$ and the target manifold $\mathcal M^{d_2,\kappa_2}$ to map onto, $GLP^{d_1,\kappa_1\to d_2,\kappa_2}_\mathbf x(\cdot)$ at $\mathbf x$ is defined as follows, 
\vspace{-0.04in}
\begin{equation}
GLP^{d_1,\kappa_1\to d_2,\kappa_2}_\mathbf x\left( \left[\begin{array}{cc}
 w & \mathbf{0}^{\top} \\
\mathbf{0} & \mathbf{W}
\end{array}\right] \right)=\left[\begin{array}{cc}
w_0 & \mathbf{0}^{\top} \\
\mathbf{0} & \mathbf{W}
\end{array}\right],
\vspace{-0.03in}
\end{equation}
so that we have 
\vspace{-0.06in}
\begin{equation}
GLP^{d_1,\kappa_1\to d_2,\kappa_2}_\mathbf x\left( \left[\begin{array}{cc}
w & \mathbf{0}^{\top} \\
\mathbf{0} & \mathbf{W}
\end{array}\right] \right)
\left[\begin{array}{c}
x_0  \\
\mathbf x_s 
\end{array}\right]=\left[\begin{array}{c}
w_0x_0  \\
\mathbf{W}\mathbf x_s 
\end{array}\right],
\vspace{-0.03in}
\end{equation}
where $w \in \mathbb R$, $w_0=\sqrt{\frac{|\kappa_1|}{|\kappa_2|}\cdot\frac{1-\kappa_2\ell(\mathbf W, \mathbf x_s)}{1-\kappa_1 \langle \mathbf x_s, \mathbf x_s\rangle}}$, and $\ell(\mathbf W, \mathbf x_s)=\left\| \mathbf W\mathbf x_s \right\|^2$.
(The derivation is given in Appendix.)


Next, we study theoretical aspects of the proposed GLP. First and foremost, we prove that GLP is \textbf{closed} in Riemannian spaces with different dimensions or curvatures, so that the mapping is done correctly.
\newtheorem*{prop3}{Proposition 2} 
\begin{prop3}
$GLP^{d_1,\kappa_1\to d_2,\kappa_2}_\mathbf x\left(\bar{ \mathbf W }\right)\mathbf x \in \mathcal M^{d_2,\kappa_2}$ holds, $\forall \mathbf x \in \mathcal M^{d_1,\kappa_1}$, where $  \bar{\mathbf{W} }=\operatorname{diag}([w, \mathbf W])$.
\end{prop3}
\vspace{-0.1in}
\begin{proof}
$\mathbf L=GLP^{d_1,\kappa_1\to d_2,\kappa_2}_\mathbf x(\bar{ \mathbf W })$, and $\langle \mathbf L\mathbf x,  \mathbf L\mathbf x \rangle_{\kappa_2} =\frac{1}{\kappa_2}$ holds.
Please refer to Appendix for the details.
\end{proof}
\noindent Second, we prove that GLP matrices cover all valid Lorentz rotation. That is, the proposed GLP can be considered as a generalization of Lorentz rotation.
\newtheorem*{prop4}{Proposition 3} 
\begin{prop4}
The set of GLP matrices projecting within $\mathcal M^{d_1, \kappa_1}$ is $\mathcal W_\mathbf x=\{\text{GLP}^{d_1,\kappa_1\to d_1,\kappa_1}_\mathbf x( \mathbf W) \}$.  Lorentz rotation set is $\mathcal Q=\{\mathbf R\}$.
$\mathcal Q \subseteq \mathcal W_\mathbf x$ holds, $\forall \mathbf x \in \mathcal M^{d_1,\kappa_1}$.
\end{prop4}
\vspace{-0.1in}
\begin{proof}
$\forall \mathbf R$, $GLP^{d_1,\kappa_1\to d_1,\kappa_1}_{\mathbf x}(\mathbf R)=\mathbf R$ holds, analog to Parseval's theorem. Please refer to Appendix for the details and further theoretical analysis.
\end{proof}

Now, we are ready to score the similarity between $\mathbf x \in \mathcal M^{d_1,\kappa_1}$ and $\mathbf y \in \mathcal M^{d_2,\kappa_2}$. 
Specifically, we add the bias for GLP, and formulate a \emph{Lorentz Layer} (LL) as follows:
\vspace{-0.05in}
\begin{equation}
\resizebox{0.885\hsize}{!}{$ 
LL^{d_1,\kappa_1\to d_2,\kappa_2}_\mathbf x\left(\mathbf W, \mathbf b, \left[\begin{array}{c}
x_0 \\
\mathbf{x}_s
\end{array}\right]  \right)
=\left[\begin{array}{c}
w_0x_0\\
 \mathbf{W}\mathbf{x}_s+\mathbf{b} 
\end{array}\right], 
$}
\vspace{-0.05in}
\end{equation}
where $\mathbf W \in \mathbb R^{d_2 \times d_1}$ and $\mathbf b \in \mathbb R^{d_2}$ denote the weight and bias, respectively. 
$\ell( \mathbf W, \mathbf x_s )=\left\| \mathbf W\mathbf x_s +\mathbf b\right\|^2$ for $w_0$.
It is easy to verify
$LL^{d_1,\kappa_1\to d_2,\kappa_2}_\mathbf x(\mathbf W,\mathbf b, \mathbf x)\in \mathcal M^{d_2,\kappa_2}$.
In this way, $\mathbf x \in \mathcal M^{d_1,\kappa_1}$ is comparable with $\mathbf y \in \mathcal M^{d_2,\kappa_2}$ after flowing over a Lorentz layer.
Accordingly, we define the generalized Lorentz similarity function as follows,
\vspace{-0.05in}
\begin{equation}
Sim^\mathcal L(\mathbf x, \mathbf y)=d_\mathcal M(LL^{d_1,\kappa_1\to d_2,\kappa_2}_\mathbf x(\mathbf W,\mathbf b, \mathbf x), \mathbf y).
\label{sim}
\end{equation}

\vspace{-0.05in}
\subsubsection{Consolidate Knowledge with Intra- \& Inter-distillation.}
In Label-free Lorentz Distillation, we jointly perform intra-distillation and inter-distillation with GLP to learn and memorize knowledge for continual graph learning, respectively.

\emph{In intra-distillation}, the student distills knowledge from the intermedian layer of itself, so that the contrastive learning is enabled without augmentation.
Specifically, we first create \emph{high-level view} and \emph{low-level view} for each node by output layer encoding and shallow layer encoding,
and then formulate the InfoNCE loss \cite{abs-1807-03748} to evaluate the agreement between different views,
\vspace{-0.12in}
\begin{equation}
\resizebox{1.02\hsize}{!}{$
\mathcal J( \mathbf x_i^{s, L},\mathbf x_i^{s, H})  =-\log \frac{\exp Sim^\mathcal L( \mathbf x_i^{s, L},\mathbf x_i^{s, H})}{\sum_{j=1}^{|\mathcal V|}\mathbb I\{i \neq j\}\exp Sim^\mathcal L( \mathbf x_i^{s, L},\mathbf x_i^{s, H})},
$}
\label{self1}
\vspace{-0.05in}
\end{equation}
where $\mathbf x_i^{s, L}$ and $\mathbf x_i^{s, H}$ denote the low-level view and high-level view of the student network, respectively. $\mathbb I\{ \cdot \} \in \{0, 1\}$ is an indicator who will return $1$ iff the condition $(\cdot)$ is true.

\emph{In inter-distillation}, the student distills knowledge from the teacher by contrasting their high-level views. We formulate teacher-student distillation objective via  InfoNCE loss,
\vspace{-0.12in}
\begin{equation}
\resizebox{1.02\hsize}{!}{$
\mathcal J(\mathbf x_i^{t, H}, \mathbf x_i^{s, H}) =-\log \frac{\exp Sim^\mathcal L(\mathbf x_i^{t, H}, \mathbf x_i^{s, H})}{\sum_{j=1}^{|\mathcal V|}\mathbb I\{i \neq j\}\exp Sim^\mathcal L(\mathbf x_i^{t, H}, \mathbf x_i^{s, H})},
$}
\vspace{-0.05in}
\label{teach1}
\end{equation}
where $\mathbf x_i^{t, H}$ and $\mathbf x_i^{s, H}$ denote the high-level view of the teacher and the student, respectively. 

Finally, with $Sim^\mathcal L(\mathbf x, \mathbf y)$ defined in Eq. (\ref{sim}), we formulate the learning objective of RieGrace as follows, 
\vspace{-0.07in}
\begin{equation}
\mathcal J_{overall} =\mathcal J_{intra}+\lambda \mathcal J_{inter},
\vspace{-0.09in}
\end{equation}
where $\lambda$ is for balance. We have contrastive loss 
$\mathcal J_{intra}=\sum\nolimits_{i=1}^{|\mathcal V|}\mathcal J(\mathbf x_i^{s, L}, \mathbf x_i^{s, H})$
and
$\mathcal J_{inter}=\sum\nolimits_{i=1}^{|\mathcal V|}\mathcal J(\mathbf x_i^{t, H}, \mathbf x_i^{s, H})$.
We summarize the overall training process of RieGrace in Algorithm 1, 
whose computational complexity is $O(|\mathcal V|^2)$ in the same order as typical contrastive models in Euclidean space, e.g., \cite{HassaniA20}.
However, RieGrace is able to consolidate knowledge of the task graph  sequence \emph{in the adaptive Riemannian spaces free of labels}.

\begin{table*}
\centering
\vspace{-0.12in}
\resizebox{1.02\linewidth}{!}{ 
\begin{tabular}{ c l | c c | c c| c c| c c| cc}
\toprule
 & \multirow{2}{*}{\textbf{Method}}  & \multicolumn{2}{c|}{\textbf{Cora}} & \multicolumn{2}{c|}{\textbf{Citeseer}} & \multicolumn{2}{c|}{\textbf{Actor}} & \multicolumn{2}{c|}{\textbf{ogbn-arXiv}} & \multicolumn{2}{c}{\textbf{Reddit}} \\
&   & PM & FM                 & PM & FM                             & PM & FM                                    & PM & FM      & PM & FM\\
\toprule
\multirow{6}{*}{\rotatebox{90}{    Euclidean   } } 
&JOINT    
& $ 93.9(0.9) $  &  $-$             &  $ 79.3(0.8)$  & $  -$                 &  $57.1(0.9)$  & $-$                      &  $82.2(0.3)$  & $-$                   & $ 96.3(0.7)$  &  $-$\\
\cline{2-12}
& ERGNN
& $ 71.1(2.5)$ & $-34.3(1.0)$  &  $65.5(0.3)$  &  $-20.4(3.9)$    &  $51.4(2.2)$   &  $-\ \ 7.2(3.2)$  &  $63.5(2.4)$  &  $-19.5(1.9)$   & $ 95.3(1.0)$ & $-23.1(1.7)$\\
& TWP
&  $81.3(3.2)$ & $-14.4(1.5)$  &  $69.8(1.5)$  & $-\ \ 8.9(2.6)$   &   $54.0(1.8)$  &  $-\ \ 2.1(1.9)$  &  $75.8(0.5)$  & $-\ \ 5.9(0.3) $  & $95.4(1.4)$ & $-\ \ \note{1.4}(1.5)$  \\
& HPN  
& $93.6(1.5)$ & $-\ \ \note{1.7}(0.7)$ & $79.0(0.9)$  & $-\ \ \note{1.5}(0.3)$    &  $56.8(1.4)$   & $-\ \ 1.5(0.9)$    & $81.2(0.7)$  &  $+\ \ \note{0.7}(0.1)$ & $95.3(0.6)$ & $-\ \ 3.6(1.0)$  \\
& FGN
& $85.5(1.4)$ & $-\ \ 2.3(1.0)$ & $73.3(0.9)$  & $-\ \ 2.2(1.7)$    &   $53.6(0.7)$   & $-\ \ 3.8(1.6)$   & $49.4(0.3)$   & $-14.8(2.2)$    & $79.0(1.8)$ & $-12.2(0.4)$  \\
& MSCGL
& $79.8(2.7)$ & $-\ \ 4.9(1.6)$ & $68.7(2.4)$  & $-\ \ 1.8(0.1)$    &   $55.9(3.3)$  & $+\ \ \note{1.3}(1.7)$   & $64.8(1.2)$   & $-\ \ 1.9(1.0)$  & $96.1(2.5)$ & $-\ \ 1.9(0.3)$  \\
& DyGRAIN
& $82.5(1.0)$ & $-\ \ 3.7(0.2)$ & $69.2(0.6)$  & $-\ \ 5.5(0.3)$    &  $56.1(1.2)$  & $-\ \ 2.9(0.3)$   & $71.9(0.2)$   & $-\ \ 4.6(0.1)$  & $93.3(0.4)$ & $-\ \ 3.1(0.2)$  \\
\midrule 
\multirow{7}{*}{\rotatebox{90}{   Riemannian }} 
& HGCN   
&  $90.6(1.8)$  & $-33.1(2.3)$  & $80.8(0.9)$  & $-21.6(0.3)$      & $56.1(1.7)$    & $-\ \ 6.3(1.6)$    & $82.0(1.5)$   & $-12.7(1.6)$    & $96.7(1.2)$ & $-33.7(0.9)$  \\
& HGCNwF
&  $88.7(2.5)$  & $-34.6(4.1)$  & $76.1(3.3)$  & $-19.9(1.5)$      & $52.8(2.9)$    & $-\ \ 8.2(2.5)$    & $78.9(2.4)$   & $-13.6(0.3)$    & $90.5(3.3)$ & $-25.0(1.7)$  \\
& LGCN
&  $91.7(0.9)$  & $-11.9(1.9)$  & $81.5(1.2)$  &$-\ \ 9.3(2.5)$  & $\note{60.2}(3.3)$   &  $-11.2(0.2)$   & $\note{82.5}(0.2)$   & $-20.8(1.1)$    & $96.1(2.4)$ & $-\ \ 9.6(2.1)$ \\
& LGCNwF 
&  $92.3(2.0)$  & $-\ \ 5.5(1.2)$& $80.3(0.7)$   & $-10.2(0.7)$     &  $57.5(1.5)$   &   $-10.9(2.4)$    & $81.3(1.8)$   & $-18.2(1.9)$   &   $95.5(0.6)$ & $-\ \ 4.9(1.5)$ \\
& $\kappa$-GCN
& $\note{93.9}(0.3)$  & $-22.0(0.4)$  & $79.8(2.9)$    & $-15.7(1.6)$     &  $56.3(3.6)$   & $-\ \ 3.1(0.9)$   & $81.6(0.3)$   & $-\ \ 9.8(1.2)$  &  $\note{96.7}(2.7)$ & $-18.6(3.3)$  \\
& $\kappa$-GCNwF
& $92.0(1.9)$  & $-11.3(2.4)$  & $\note{81.0}(0.5)$  & $-\ \ 6.1(1.2)$   &  $59.7(2.0)$   & $+\ \ 0.6(0.3)$  & $79.9(1.9)$   & $-\ \ 5.1(2.0)$  &  $94.1(1.0)$ & $-11.5(2.4)$  \\
\cline{2-12}
& \textbf{RieGrace}
& $\mathbf{95.2}(0.8)$ & $-\ \ \mathbf{1.2}(0.7)$&$\mathbf{83.6}(2.4)$ &$-\ \ \mathbf{1.3}(0.6)$  &   $\mathbf{61.9}(1.2)$ & $+\ \ \mathbf{1.9}(1.1)$  & $\mathbf{83.9}(0.3)$ & $+\ \ \mathbf{1.2}(0.5)$ &   $\mathbf{97.9}(1.8)$ & $-\ \ \mathbf{1.1}(1.5)$  \\
\bottomrule
\end{tabular} }
\vspace{-0.09in}
\caption{Node classification on Citerseer, Cora, Actor, ogbn-arXiv and Riddit. We report both PM(\%) and FM(\%). Confidence interval is given in brackets. The best scores are in \textbf{bold}, and the second \underline{underlined}. }
\vspace{-0.15in}
\label{results}
\end{table*}

\vspace{-0.1in}
\section{Experiment}
\vspace{-0.05in}
We conduct extensive experiments  on a variety of datasets with the aim to answer following research questions (\emph{RQs}):
\vspace{-0.02in}
\begin{itemize}
  \item \textbf{\emph{RQ1}}: How does the proposed \emph{RieGrace} perform?
  \item \textbf{\emph{RQ2}}: How does the proposed component, either \emph{CurvNet} or \emph{GLP}, contributes to the success of RieGrace?
    \item \textbf{\emph{RQ3}}: How does the \emph{curvature} change over the graph sequence in continual learning?
\end{itemize}
\vspace{-0.11in}
\subsubsection{Datasets.} 
We choose five benchmark datasets, i.e.,
\textbf{Cora} and
\textbf{Citeseer} \cite{DBLP:journals/aim/SenNBGGE08}, 
\textbf{Actor} \cite{DBLP:conf/kdd/TangSWY09},
\textbf{ogbn-arXiv} \cite{DBLP:conf/nips/MikolovSCCD13}
and
\textbf{Reddit} \cite{hamilton2017inductive}.
The setting of graph sequence (task continuum) on Cora, Citerseer, Actor and ogbn-arXiv follows \citet{DBLP:journals/corr/abs-2111-15422},  and the setting on Reddit follows \citet{DBLP:conf/aaai/0002C21}.

\vspace{-0.07in}
\subsubsection{Euclidean Baseline.} 
We choose several strong baselines, i.e., \textbf{ERGNN} \cite{DBLP:conf/aaai/0002C21}, \textbf{TWP} \cite{DBLP:conf/aaai/LiuYW21}, \textbf{HPN} \cite{DBLP:journals/corr/abs-2111-15422}, \textbf{FGN} \cite{FGN}, \textbf{MSCGL} \cite{MCGL-NAS} and \textbf{DyGRAIN} \cite{DBLP:conf/ijcai/KimYK22}.
ERGNN, TWP and DyGRAIN are implemented with GAT backbone \cite{velickovic2018graph}, which generally achieves the best results as reported.
We also include the joint training with GAT (\textbf{JOINT}) that trains all the tasks jointly. \emph{Since no catastrophic forgetting exists, it approximates the upper bound in Euclidean space w.r.t. GAT}. 
MSCGL is designed for multimodal graphs, and we use the corresponding unimodal version to fit the benchmarks.
Existing methods are trained in supervised fashion, and we propose the first self-supervised model for continual graph learning to our knowledge.

\vspace{-0.07in}
\subsubsection{Riemannian Baseline.} 
In the literature, there is no continual graph learner in Riemannian space. 
Alternatively, we fine-tune the offline Riemannian GNNs in each learning session, in order to show the forgetting of continual learning in Riemannian space. Specifically, we choose \textbf{HGCN} \cite{HGCN},
\textbf{LGCN} \cite{ZhangWSLS21},
and
$\mathbf \kappa$-\textbf{GCN} \cite{BachmannBG20}.
In addition, we implement the supervised LwF \cite{DBLP:journals/pami/LiH18a} for CNNs on these Riemannian GNNs (denoted by -\textbf{wF} suffix), in order to show adapting existing methods to Riemannian GNNs trends to result in inferior performance.

\vspace{-0.07in}
\subsubsection{Evaluation Metric.}
Following \citet{MCGL-NAS,DBLP:conf/aaai/0002C21,DBLP:conf/nips/Lopez-PazR17}, we utilize Performance Mean (PM) and Forgetting Mean (FM) to measure the learning and memorizing abilities, respectively. 
Negative FM means the existence of forgetting, and positive FM indicates positive knowledge transfer between tasks.

\vspace{-0.07in}
\subsubsection{Euclidean Input.} 
The input feature $\mathbf x$ are Euclidean by default.
To bridge this gap, we formulate an input transformation for Riemannian models, $\Gamma_\kappa: \mathbb R^d \to \mathcal M^{d, \kappa}$. Specifically, we have $\Gamma_\kappa(\mathbf x)=exp_\mathcal O^\kappa([0||\mathbf x])$, and $\kappa$ is either given by CurNet in RieGrace, or set as a parameter in other models.

\vspace{-0.07in}
\subsubsection{Model Configuration.}
In our model, we stack the convolutional layer twice with a 2-layer CurvNet. Balance weight $\lambda=1$. 
As a self-supervised model, 
RieGrace first learns encodings without labels, and then the encodings are directly utilized for training and testing, similar to \citet{VelickovicFHLBH19}. 
The grid search is performed for hyperparameters, e.g., learning rate: $[0.001, 0.005, 0.008, 0.01]$.

\noindent (\emph{Appendix} gives the details on datasets, baselines, metrics, implementation as well as the further mathematics.)

\vspace{-0.1in}
\subsection{Main Results (\emph{RQ1})}
\vspace{-0.02in}

Node classification is utilized as the learning task for the evaluation.
Traditional classifiers work with Euclidean space, and cannot be applied to Riemannian spaces due to the essential distinction in geometry. 
For Riemannian methods, we extend the classification method proposed in \citet{HGNN} to Riemannian space of arbitrary curvature with distance metric $d_\mathcal M$ given in Table \ref{tab:ops}.
For fair comparisons, we perform $10$ independent runs for each model, and report the mean value with $95\%$ confidence interval in Table \ref{results}. 
Dimension is set to $16$ for Riemannian models, and follows original settings for Euclidean models.
As shown in Table 2, traditional continual learning methods suffers from forgetting in general, though MSCGL, HPN, $\kappa$-GCNwF and our RieGrace have positive knowledge transfer in a few cases.
Our self-supervised RieGrace achieves the best results in both PM and FM, even outperforming the supervised models.
The reason is two-fold:
i) RieGrace successively matches each task graph with adaptive Riemannian spaces, improving the learning ability.
ii) RieGrace learns from the teacher to preserve past knowledge in the label-free Lorentz distillation, improving the memorizing ability.

\begin{table}
\vspace{-0.01in}
\centering
\resizebox{0.91\linewidth}{!}{ 
\begin{tabular}{ c | c c| c c }
\toprule
   \multirow{2}{*}{\textbf{Variant}}  & \multicolumn{2}{c|}{ \textbf{Citeseer}} & \multicolumn{2}{c}{\textbf{Actor} }   \\
   & PM & FM                 & PM & FM                             \\
\toprule
$\mathbb{S}$w/oL
&  $66.7(0.3)$   & $-\ \ 6.7(0.9) $   &  $51.6(0.8)$   &  $-7.1(0.7) $    \\
$\mathbb{S}$
&  $70.2(1.5)$   & $-\ \ 5.3(1.0) $   &  $53.4(3.1)$   &  $-0.9(0.2) $    \\
\midrule 
$\mathbb{E}$
&  $69.8(0.9)$   & $-11.9(0.3)$      &  $52.9(2.7)$   & $-4.3(1.6)$  \\
\midrule 
$\mathbb{H}$w/oL
&  $77.1(3.5)$   & $-\ \ 8.2(0.8)$      &  $53.3(1.5)$   & $-8.9(0.7)$  \\
$\mathbb{H}$
&  $80.9(0.2)$   & $-\ \ 5.7(2.1)$      &  $56.6(2.4)$   & $-4.8(0.1)$  \\
\midrule 
$\mathcal{M}$w/oL
&  $\note{81.2}(1.8)$ & $-\ \ \note{3.9}(2.2)$  &  $\note{58.5}(0.6)$  & $+\note{0.5}(1.3)$  \\
\textbf{Full}
&  $\mathbf{83.6}(2.4)$ &$-\ \ \mathbf{1.3}(0.6)$  &   $\mathbf{61.9}(1.2)$ & $+ \mathbf{1.9}(1.1)$   \\
\bottomrule
\end{tabular} }
\vspace{-0.09in}
\caption{Ablation study on Citersser and Actor. Confidence interval is given in bracket. The best scores are in \textbf{bold}.}
\vspace{-0.24in}
\label{ablation}
\end{table}

\vspace{-0.10in}
\subsection{Ablation Study (\emph{RQ2})}
\vspace{-0.02in}
We conduct ablation study to show how each proposed component contributes to the success of  RieGrace.
To this end, we design two kinds of variants described as follows:

\noindent{\emph{i) To verify the importance of GLP directly mapping between Riemannian spaces,}}  we design the variants that involve a tangent space for the mapping, denoted by -w/oL suffix. Specifically, we replace the Lorentz layer by logarithmic and exponential maps in corresponding models. 

\noindent{\emph{ii) To verify the importance of CurvNet supporting curvature adaptation to any positively or negatively curved spaces,}} 
we design the variants restricted in hyperbolic, Euclidean and hyperspherical space, denoted by $\mathbb S$, $\mathbb E$ and $\mathbb H$.
Specifically, we replace CurvNet by the parameter $\kappa$ of the given sign, and we use corresponding Euclidean operators for $\mathbb E$ variant.

We have six variants in total. We report their PM and FM in Table \ref{ablation}, and find that:
i)  The proposed RieGrace with GLP beats the tangent space-based variants. It suggests that introducing an additional tangent space weakens the performance for contrastive distillation.
ii) The proposed RieGrace with CurvNet outperforms constrained-space variants ($\mathbb S$, $\mathbb E$ or $\mathbb H$). We will give further discussion in the case study.

\vspace{-0.07in}
\subsection{Case Study and Discussion (\emph{RQ3})}
\vspace{-0.02in}

We conduct a case study on \textbf{ogbn-arXiv} to investigate on the curvature over the graph sequence in continual learning.

We begin with evaluating the effectiveness of CurvNet.
To this end, we leverage the metric of embedding distortion, which is minimized with proper curvature \cite{DBLP:conf/icml/SalaSGR18}.
Specifically, given an embedding $\Psi: v_i \in \mathcal V \to \mathbf x_i \in \mathcal M^{d,\kappa}$ on a graph $G$, 
the embedding distortion is defined as
$D_{G, \mathcal M}=\frac{1}{|\mathcal V|^2} \sum\nolimits_{i,j \in \mathcal V}\left| 1-\frac{d_\mathcal M(\mathbf x_i, \mathbf x_j)}{d_G(v_i, v_j)}\right| $,
where $d_\mathcal M(\mathbf x_i, \mathbf x_j)$ and $d_G(v_i, v_j)$ denote embedding distance and graph distance, respectively.
Graph distance $d_G(v_i, v_j)$ is defined on the shortest path between $v_i$ and $v_j$ regarding $d_\mathcal M$, e.g., if the shortest path between $v_A$ and $v_B$ is $v_A \to v_C \to v_B$, then we have $d_G(v_A, v_B)=d_\mathcal M(\mathbf x_A, \mathbf x_C)+d_\mathcal M(\mathbf x_C, \mathbf x_B)$.
We compare CurvNet with the combinational method proposed in  \cite{BachmannBG20}, termed as ComC. 
We report the distortion $D_{G, \mathcal M}$  in $16$-dimensional Riemannian spaces with the curvature estimated by CurvNet and ComC in Table \ref{curvature}, where $D_{G, \mathcal M}$ of  $128$-dimensional Eulidean space is also listed (ZeroC).
As shown in Table \ref{curvature}, our CurvNet gives a better curvature estimation than ComC, and ZeroC results in larger distortion even with high dimension.

Next, we estimate the curvature over the graph sequence via CurvNet, which is jointly learned with RieGrace.
We illustrate the shape of Riemannian space with corresponding curvatures with a $2$-dimensional visualization on ogbn-arXiv in Figure 2.
As shown in Figure 2, rather than remains in a certain type of space, \emph{the underlying geometry varies from positively curved hyperspherical spaces to negatively curved hyperbolic spaces in the graph sequence}. 
It suggests the necessity of curvature adaptation supporting the shift among any positive and negative values. 
The observation in both Table \ref{curvature} and Figure \ref{manifold} motivates our study indeed, and essentially explains the inferior of existing Euclidean methods and the superior of our RieGrace.

\begin{table}
\centering
\vspace{-0.05in}
\resizebox{0.9\linewidth}{!}{ 
\begin{tabular}{ c | c c c }
\toprule
 $D_{G, \mathcal M}$   & Task Graph 1 & Task Graph 2   & Task Graph 3                          \\
\toprule
  CurvNet &  $ \mathbf{0.435} (0.027) $  &  $\mathbf{0.490}(0.010) $  &  $\mathbf{0.367}(0.082)$   \\
 ComC     &  $ 0.507 (0.012) $  &  $0.653 (0.007) $   &  $ 0.524(0.033)$   \\
 ZeroC     &  $ 5.118(0.129)  $   &  $3.967(0.022) $    & $ 4.025(0.105)$  \\
\bottomrule
\end{tabular} 
}
\vspace{-0.05in}
\caption{Embedding distortion $D_{G, \mathcal M}$ with different curvatures on ogbn-arXiv. Confidence interval is given in bracket.}
\vspace{-0.2in}
\label{curvature}
\end{table}

\begin{figure} 
 \vspace{-0.1in}
\centering 
\resizebox{1.02\linewidth}{!}{ 
\subfigure[$G_1$, $\kappa=0.227$]{
\includegraphics[width=0.325\linewidth]{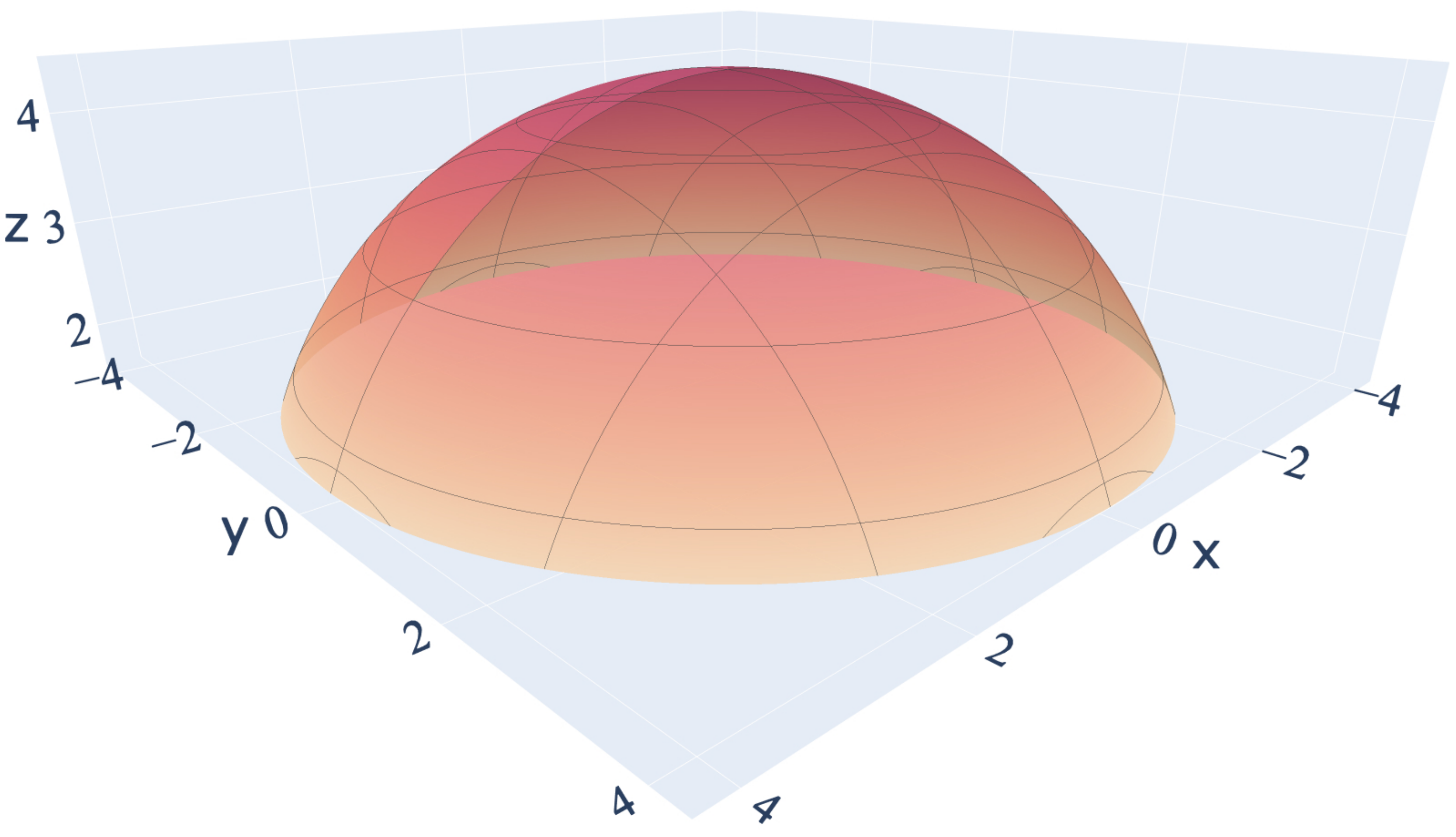}}
\hspace{-0.025\linewidth}
\subfigure[$G_2$, $\kappa=-0.536$]{
\includegraphics[width=0.325\linewidth]{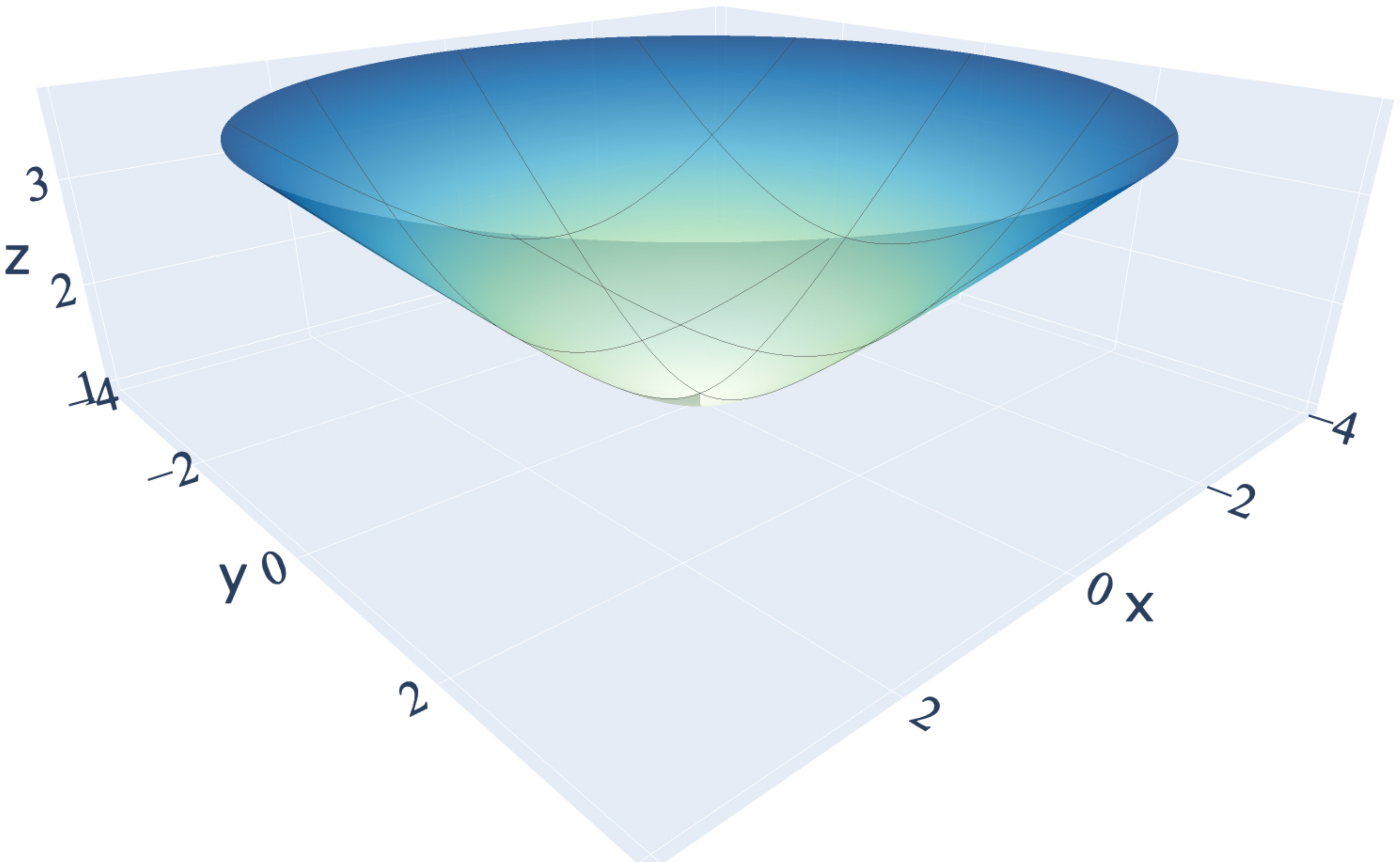}}
\hspace{-0.025\linewidth}
\subfigure[$G_3$, $\kappa=-1.073$]{
\includegraphics[width=0.325\linewidth]{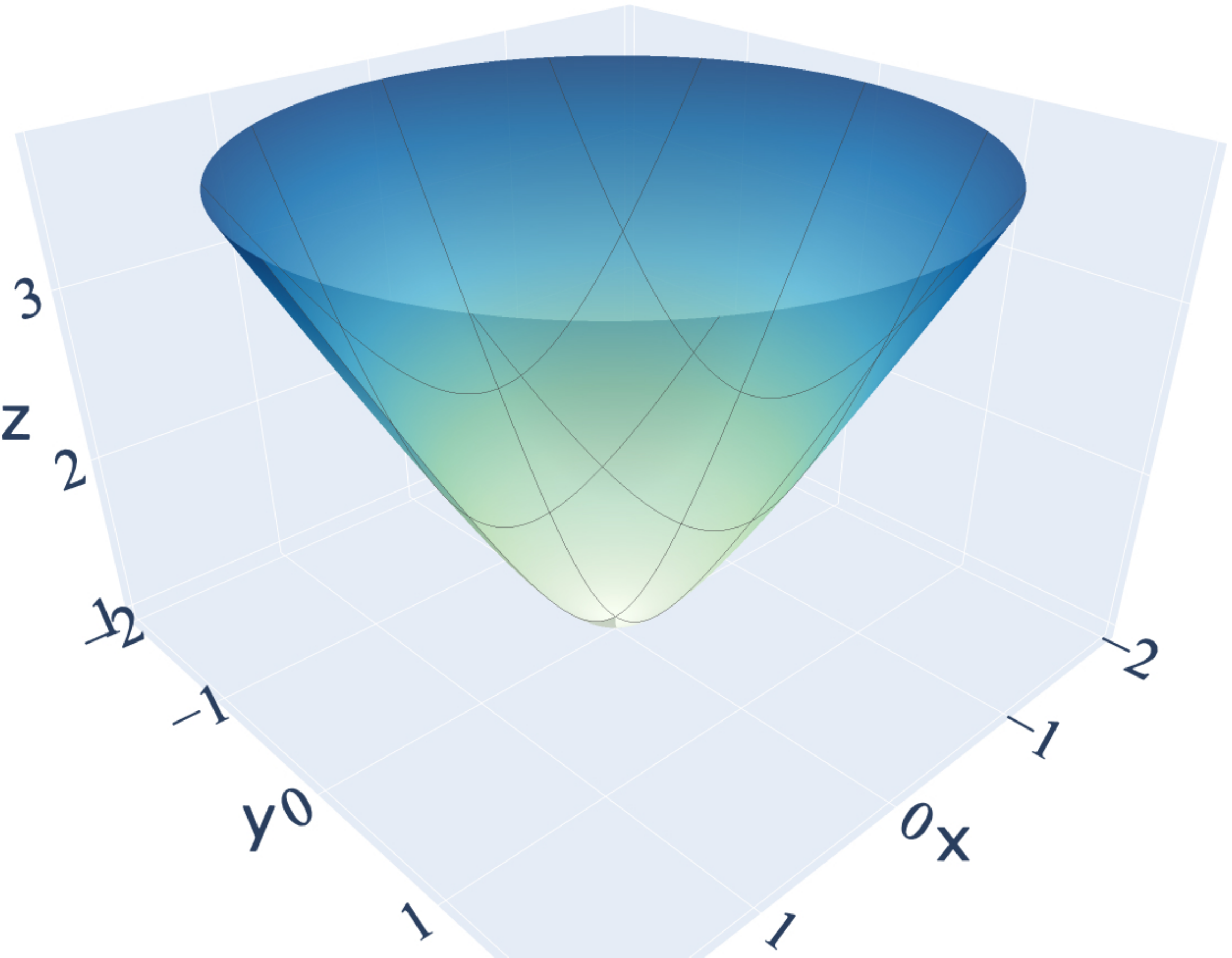}}
}
 \vspace{-0.18in}
\caption{Illustration of the Riemannian spaces in the task graphs $G_t$ on ogbn-arXiv. $\kappa$ is the learnt curvature.}
 \vspace{-0.22in}
\label{manifold}
\end{figure}

\vspace{-0.1in}
\section{Related Work}
\vspace{-0.03in}
\subsubsection{Continual Graph Learning.}
Existing studies can be roughly divided into three categories, i.e., replay (or rehearsal), regularization and architectural methods \cite{DBLP:journals/corr/abs-2202-10688}.
Replay methods retrain representative samples in  the memory or pseudo-samples to survive from catastrophic forgetting, e.g.,
ERGNN \cite{DBLP:conf/aaai/0002C21} introduces a well-designed strategy to select the samples.
HPN \cite{DBLP:journals/corr/abs-2111-15422} extends knowledge with the prototypes learnt from old tasks.
Regularization methods append a regular term to the loss to preserve the utmost past knowledge, e.g.,
TWP \cite{DBLP:conf/aaai/LiuYW21} preserves important parameters for both task-related and topology-related goals.
MSCGL \cite{MCGL-NAS} is designed for multimodal graphs with neural architectural search.
DyGRAIN \cite{DBLP:conf/ijcai/KimYK22} explores the adaptation of receptive fields while distilling knowledge.
Architectural methods modify the neural architecture of graph model itself, such as FGN \cite{FGN}.
Meanwhile, continual graph learning has been applied to recommendation system \cite{DBLP:conf/cikm/XuZGGTC20}, trafficflow prediction \cite{DBLP:conf/ijcai/ChenWX21}, etc.
In addition, \citet{DBLP:conf/cikm/WangSWW20} mainly focus on a related but different problem with the time-incremental setting.
Recently,  \citet{DBLP:conf/wsdm/TanDG022,DBLP:journals/corr/abs-2205-13954} study the few-shot class-incremental learning on graphs which owns essentially different setting to ours.
Since no existing work is suitable for the self-supervised continual graph learning, we are devoted to bridging this gap in this work.

\vspace{-0.05in}
\subsubsection{Riemannian Representation Learning.}
It has achieved great success in a variety of applications \cite{mathieu2019continuous,HAN,nagano2019wrapped,DBLP:conf/icdm/0008Z0WDSY20}.
Here, we focus on Riemannian models on graphs.
In hyperbolic space,
\citet{nickel2017poincare,suzuki2019hyperbolic} introduce shallow models,
while HGCN \cite{HGCN}, HGNN \cite{HGNN} and LGNN \cite{ZhangWSLS21} generalize convolutional network with different formalism under static setting.
Recently, HVGNN \cite{HVGNN} and HTGN \cite{HTGN} extend hyperbolic graph neural network to temporal graphs. 
Beyond hyperbolic space,
\citet{DBLP:conf/icml/SalaSGR18} study the matrix manifold of Riemannian spaces.
$\kappa$-GCN \cite{BachmannBG20} extends GCN to constant-curvature spaces with $\kappa$-sterographical model, but its formalism cannot be applied to our problem.
\citet{DBLP:conf/www/YangCPLYX22} model the graph in the dual space of Euclidean and hyperbolic ones.
\citet{GuSGR19}  and \citet{DBLP:conf/www/WangWSWNAXYC21} explore the mixed-curvature spaces, and \citet{SelfMGNN} propose the first self-supervised GNN in mixed-curvature spaces.
\citet{NEURIPS2021_b91b1fac} and \citet{DBLP:conf/kdd/XiongZNXP0S22} study graph learning on a kind of pseudo Riemannian manifold, ultrahyperbolic space.
Recently, \citet{DBLP:conf/cikm/0008YPY22} propose a novel GNN in general on Riemannian manifolds with the time-varying curvature.
All existing Riemannian models adopt offline training, and we propose the first continual graph learner in Riemannian space to the best of our knowledge.

\vspace{-0.07in}
\section{Conclusion}
\vspace{-0.02in}
In this paper, we propose the first self-supervised continual graph learner in adaptive Riemannian spaces, RieGrace.
Specifically, we first formulate a unified GNN coupled with the CurvNet, so that Riemannian space is shaped by the learnt curvature adaptive to each task graph.
Then, we propose Label-free Lorentz Distillation approach to consolidate knowledge without catastrophic forgetting,
where we perform contrastive distillation in Riemannian spaces with the proposed GLP.
Extensive experiments on the benchmark datasets show the superiority of RieGrace.

\vspace{-0.05in}

\section{Acknowledgments}
The authors of this paper were supported in part by National Natural Science Foundation of China under Grant 62202164, the National Key R\&D Program of China through grant 2021YFB1714800,  S\&T Program of Hebei through grant 21340301D and the Fundamental Research Funds for the Central Universities 2022MS018.
Prof. Philip S. Yu is supported in part by NSF under grants III-1763325, III-1909323,  III-2106758, and SaTC-1930941.
Corresponding Authors: Li Sun and Hao Peng.

\bibliography{aaai23}

\bigskip

\end{document}